\documentclass{article} % For LaTeX2e
\usepackage{mytemplate,times}

\usepackage{hyperref}
\usepackage{url}
\usepackage{algorithm}
\usepackage{algorithmic}
\usepackage{graphicx}
\usepackage{subfigure}

\title{Joint action loss for proximal policy optimization}

\author {
    Xiulei Song\thanks{Contributed equally} \\
    JumpW AI Group\\
    Shanghai,China\\
    \texttt{songxiulei@jumpw.com}\\
    \And
    Yizhao Jin\footnotemark[1]\quad \& Greg Slabaugh \& Simon Lucas\\
    Queen Mary University of London Game AI Group\\
    London, United Kingdom\\
    \texttt{\{acw596, g.slabaugh, simon.luacs\}qmul@email}\\
}

\begin{document}
\maketitle
\begin{abstract}
PPO (Proximal Policy Optimization) is a state-of-the-art policy gradient algorithm that has been successfully applied to complex computer games such as Dota 2 and Honor of Kings. In these environments, an agent makes compound actions consisting of multiple sub-actions. PPO uses clipping to restrict policy updates. Although clipping is simple and effective, it is not efficient in its sample use. For compound actions, most PPO implementations consider the joint probability (density) of sub-actions, which means that if the ratio of a sample (state compound-action pair) exceeds the range, the gradient the sample produces is zero. Instead, for each sub-action we calculate the loss separately, which is less prone to clipping during updates thereby making better use of samples. Further, we propose a multi-action mixed loss that combines joint and separate probabilities. We perform experiments in Gym-$\mu$RTS and MuJoCo. Our hybrid model improves performance by more than 50\% in different MuJoCo environments compared to OpenAI's PPO benchmark results. And in Gym-$\mu$RTS, we find the sub-action loss outperforms the standard PPO approach, especially when the clip range is large. Our findings suggest this method can better balance the use-efficiency and quality of samples.
\end{abstract}

\noindent PPO~\cite{schulman2017proximal} is a simple and efficient reinforcement learning algorithm inspired by TRPO (Trust Region Policy Optimization)~\cite{schulman2015trust}. A major challenge for PPO and other policy gradient algorithms is to estimate the correct step of policy updates, since input data is heavily dependent on the current policy~\cite{schulman2015trust}. TRPO solves this problem by imposing a trust domain constraint on the target function to control the KL divergence between the old and new policies. PPO significantly reduces complexity by employing a clipping mechanism, which attempts to eliminate incentives that push the policy away from the old policy when the likelihood ratio between the two exceeds a limit. 

There are existing studies that examine the clipping mechanism of PPO. Engstrom et al.~\cite{engstrom2020implementation} conducted an experiment on whether a clip mechanism was needed in PPO. In their experiment, PPO code-level optimizations played a greater role than the clipping mechanism. Separately, Wang et al.~\cite{wang2020truly} used a new clipping function to support rollback behavior to limit differences between new and old policies. The trigger condition of clipping was replaced by the condition based on the trust region, so that the agent's objective function generated by optimization can guarantee monotonic improvement of the final policy performance.

PPO has been successfully used in some complex games, such as Dota 2~\cite{berner2019dota} and Glory of Kings~\cite{ye2020mastering}. The actions of these games' agent are compound actions that consist of several sub-actions. A common treatment of compound actions in the execution of PPO is based on the probability (or probability density) of the entire action, as shown in Eq.~\ref{r1}.  However, one problem with this approach is that for compound actions, clipping will filter out the entire action data that does not meet the requirements, even if the compound action can be adjusted in the dimensions of some sub-actions. From another point of view, it is feasible to calculate the loss separately for each sub-action. A problem with considering each sub-action individually is that it does not consider the connections between the sub-actions. We address this by introducing a simple mixing of both approaches in this paper.

In the next part of the paper, the PPO algorithm will be introduced, with its approach to clipping as the focus of our research. Then, we analyze PPO for compound action environments and present our novel PPO loss functions. We conduct experiments with these loss functions in Gym-$\mu$RTS~\cite{huang2021gym} and MuJoCo~\cite{todorov2012MuJoCo}, which are environments for discrete compound actions and continuous compound actions, respectively. In addition to the common loss function for the entire combined action, we also study the loss function of each sub-action and the mixed loss function of the two.

\section{Proximal Policy Optimization}
The mathematical basis of the main idea of TRPO and PPO is importance sampling: $x_i$ is sampled from the policy distribution $p(x)$. However, the value of $p(x)$ is often not directly available, and must be obtained through other distributions, namely the old policy $q(x)$ obtained by indirect sampling.
\begin{equation}
    E_{x\sim p} [p(x)f(x)]=E_{x\sim q} [\frac{p(x)}{q(x)}f(x)]
\label{importence_sample}
\end{equation}
%\end{center}
The distributions $p(x)$ and $q(x)$ should be similar in order to achieve good results. Specifically, because the new strategy distribution in the original formula of the strategy gradient is difficult to obtain, the old strategy is used for indirect sampling. This results in the ratio $r_{t}(\theta)$ of PPO, expressed as 
\begin{equation}
    r_{t}(\theta) = \frac{\pi_{\theta}(a_t|s_t)}{\pi_{\theta_{old}}(a_t|s_t)}
\label{ratio}
\end{equation}
where $\pi_{\theta}$ is the new policy, $\pi_{\theta_{old}}$ is the sample policy, $a_t$ is the action at the time of sampling, and $s_t$ is the state at the time of sampling.  PPO~\cite{schulman2017proximal} is restricted by the clipping mechanism, whereas TRPO~\cite{schulman2015trust} restricts the policy by explicitly imposing constraints, and Wang et al.~\cite{wang2020truly} used a new clipping method to soft clip the ratio. In general, these methods are designed to enforce the trust region. If $\pi_{\theta}(a_t|s_t)$ and $\pi_{\theta_{old}}(a_t|s_t)$ are very different, this is not conducive to convergence. PPO uses clipping to cull out-of-range data. If the action is a good action, the PPO restricts it from over-updating in a favorable direction (probability increases), and if it is a bad action, it restricts it from over-updating in an unfavorable direction (probability decreases), so as to limit the magnitude of the policy update.
\begin{equation}
    L^{CLIP} = min(r_{t}\hat{A}_{t},clip(r_{t}, 1-\epsilon,1+\epsilon)\hat{A}_{t})
\label{lclip}
\end{equation}
where $\hat{A}_{t}$ is an estimator of the advantage function at time $t$ and $\epsilon$ is a hyperparameter.
Eq.~\ref{lclip} can be written as follows in Eq.~\ref{lclip_g}. Once the ratio exceeds the clipping range, the gradient of the update strategy will be zero. In other words, data beyond the clipping range will be filtered out.
\begin{equation}
    L^{CLIP} = \left\{ 
    \begin{array}{lllll}
     (1-\epsilon)\hat{A}_{t}&  &r_{t}\leq1-\epsilon& and &\hat{A}_{t}<0\\
     (1+\epsilon)\hat{A}_{t}&  &r_{t}>1+\epsilon& and &\hat{A}_{t}>0\\
     r_{t}\hat{A}_{t}&  &otherwise&&
    \end{array}
    \right.
\label{lclip_g}
\end{equation}
The overall loss $J(\theta)$ of PPO consists of three parts: policy loss $L^{CLIP}$, value loss $L^{VT}$, and entropy loss $S_{\pi_{\theta}}$,
\begin{equation}
    L^{VT} = (V_{\theta}(s_t) - V_{t} ^{target})^2
\label{lvalue}
\end{equation}
\begin{equation}
    S_{\pi_{\theta}}=\sum -p_{a}log(p_{a})
\label{entropy}
\end{equation}
\begin{equation}
    J(\theta) = {E}_{t}[L^{CLIP}-c_{1}L^{VT}+c_{2}S_{\pi_{\theta}}]
\label{loss}
\end{equation}
where $V$ is the value, $c_{1}$ is value loss coefficient, and $c_{2}$ is entropy loss coefficient.

\section{Compound Action}
%more explain of compound action and sub-action
In many environments, the action of an agent is a compound action that is composed of multiple sub-actions. For example, often in real-time strategy (RTS) games, the action interacting with the environment is composed of a selected unit, action type, action direction and action target. And when controlling some robots, an action consists of the movement of each joint.

The optimization objective of the policy gradient can be expressed as Eq.~\ref{goal}. When a certain state action pair $(s, a)$ in the trajectory is satisfactory, that is, $q(s, a)>0$, increase the probability of $\pi_{\theta}(s, a)$, and vice versa.
%\begin{center}
\begin{equation}
%$J_{\theta} = {E}_{\pi_{\theta}}\left[{\Psi_t\log \pi_{\theta}(a_t|s_t)} \right]
    J_{\theta} = E_{\pi_{\theta}}\left[{\Psi_t\log \pi_{\theta}(a_t|s_t)} \right]
\label{goal}
\end{equation}
%\end{center}
For the compound action environment, action $a_{t}$ in Eq.~\ref{goal} is the entire compound action. Assuming that the sub-actions of the composite action are independent of each other, then $\pi_{\theta}(a_{t}|s_{t}) = \prod\pi^{i}(a^{i}_{t}|s_{t})$, where $a^{i}_{t}$ is the $i$th sub-action. 
%\begin{center}
\begin{equation}
    J^{i}_{\theta} = {E}_{\pi_{\theta}}\left[{\Psi_{t}\log \pi^{i}_{\theta}(a^{i}_{t}|s_t)}\right]
\label{sub_action_goal}
\end{equation}
%\end{center}
%Assuming that sub-actions are independent of each other,
Under the same independence assumption,
$\log \pi_{\theta}(a_t|s_t) = \sum_{i}\log\pi^{i}(a^{i}_{t}|s_{t})$, so
%\begin{center}
\begin{equation}
{E}_{\pi_{\theta}}\left[{\Psi_{t}\sum_{i}\log \pi^{i}_{\theta}(a^{i}_{t}|s_t)} \right] = \sum_{i}{E}_{\pi_{\theta}}\left[{\Psi_{t}\log \pi^{i}_{\theta}(a^{i}_{t}|s_t)} \right]
\end{equation}
\begin{equation}
    J_{\theta} =   \sum_{i}J^{i}_{\theta}
\label{jeq}
\end{equation}
%\end{center}
But the situation is different for PPO. We will discuss this in detail in the following subsection.
\subsection{Compound ratio}
For compound actions, a natural choice is to multiply the probabilities (or probability densities) of the sub-actions to obtain the probability (or probability density) of the compound action, thus obtaining the compound ratio (Eq.~\ref{r1}). This is a common method for PPO implementation~\cite{schulman2017proximal,huang2021gym,engstrom2020implementation,wang2020truly,wang2019trust}. 
%\begin{center}
\begin{equation}
    r_{1} = \frac{\prod\pi^{i}_{new}(a^{i}_{t}|s_{t})}{\prod\pi^{i}_{old}(a^{i}_{t}|s_{t})}
\label{r1}
\end{equation}
%\end{center}
Without considering clipping, the optimization objective is shown in the following equation (\ref{ppogoal}).
%\begin{center}
\begin{equation}
    J_{\theta} = {E}_{\pi_{\theta}}\left[\hat{A}_t\frac{\prod\pi^{i}_{new}(a^{i}_{t}|s_{t})}{\prod\pi^{i}_{old}(a^{i}_{t}|s_{t})} \right] 
\label{ppogoal}
\end{equation}
%\end{center}

\subsection{Sub-action ratio}
\begin{figure*}[t]
    \centering
    \includegraphics[width=0.8\textwidth]{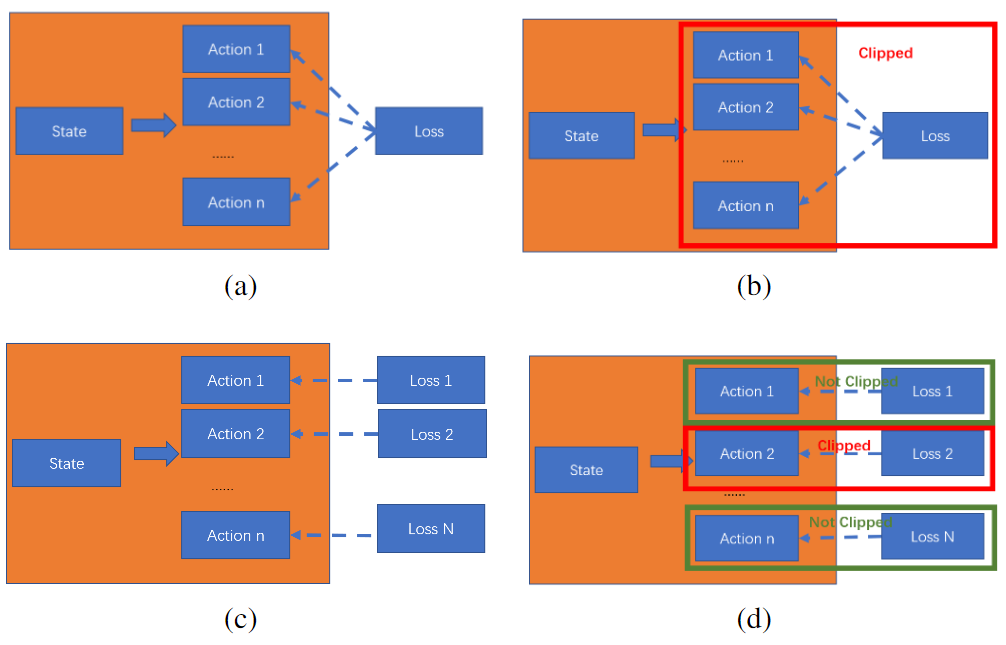}
    \caption{This figure shows the difference between the compound action loss and the sub-action loss. If the compound action loss is used, as shown in (a), there is one loss for the entire compound action to update the policy. With the sub-action loss, shown in (c), there is a loss for each sub-action in a composite action to update the policy. When using the compound action loss, if the ratio is out of range as shown in (b), the entire likelihood ratio is clipped. However with sub-action loss as shown in (d), for one sample, it is possible that part of the ratio will be clipped (red box) and the other parts will not be clipped (green boxes). This means that less clipping occurs, which improves the efficiency of sample use.}
    \label{compoundactionandsub-action}
\end{figure*}

An improvement of PPO is to change the KL divergence in the policy loss function to clip on the basis of TRPO to constrain the range of the ratio and remove the data generated by the policy with too large a gap. This method requires less computation than TRPO, but it will roughly filter out data that exceeds the clip range, which reduces the efficiency of sample usage. A simple way to improve the pass rate of data is to expand the range of clipping, but this will result in unnecessary data passing through the filter. Alternatively, one can  calculate the ratio for each sub-action $a^{i}_{t}$, as shown in Eq.~\ref{r2}. With the compound ratio (Eq.~\ref{r1}), the algorithm's optimization goal is the entire compound action, and with the sub-action ratio of Eq.~\ref{r2}, the algorithm will optimize each sub-action. If each sub-action is optimized then the entire compound action is also optimized. In subsequent experiments, we verified that this method is feasible. As shown in Figures \ref{compoundactionandsub-action} and \ref{unclip_number}, more samples will be unclipped with sub-action-loss.

%\begin{center}
\begin{equation}
    r^{i}_{2} = \frac{\pi^{i}_{new}(a^{i}_{t}|s^{i}_{t})}{\pi^{i}_{old}(a^{i}_{t}|s^{i}_{t})}
\label{r2}
\end{equation}
%\end{center}

Without considering clipping, the optimization objective is shown in the following equation (Eq.~\ref{pposubactiongoal}). Obviously, Eqns.~\ref{ppogoal} and \ref{pposubactiongoal} cannot be the same as Eq.~\ref{jeq}. 

%\begin{center}
\begin{equation}
    J_{\theta} = {E}_{\pi_{\theta}}\left[\hat{A}_t\sum\frac{\pi^{i}_{new}(a^{i}_{t}|s_{t})}{\pi^{i}_{old}(a^{i}_{t}|s_{t})} \right] 
\label{pposubactiongoal}
\end{equation}
%\end{center}

In the case of clipping, for Eq.~\ref{r1}, if the ratio of compound action exceeds the clipping range, the entire data of this action will be ignored. If Eq.~\ref{r2} is adopted, each sub-action is clipped individually, and for a compound action, part of its sub-actions are filtered out and the rest will be optimized for the strategy. As mentioned in Section 2, the effect of clipping is that if the action is a good action, PPO restricts it from over-updating in a favorable direction, and if it is an unfavorable action, it restricts it from over-updating in the unfavorable direction. Compared with Eqns.~\ref{r1} and~\ref{r2} provides more effective data to adjust the sub-action. In the case where there is not much data filtered out by clipping, the policy can be more quickly optimized.

\subsection{Mixed loss}
 We propose a mixed method that can be used to reduce the impact of excessively large ratios while increasing the sample efficiency. Combining equations \ref{lclip}, \ref{r1} and \ref{r2}, the mixed ratio represented by Eqn.~\ref{mix_ratio} Eq.~\ref{mix_ratio_loss} can be obtained, where $w$ is used to weight the two ratios. We suppose that the two ratios work similarly; while the optimal weight could be found for a particular environment, we did not deeply explore this hyper-parameter. In the experiments in this paper, the two ratios have the same weight, that in Eq.~\ref{mix_ratio_loss} with $w=0.5$, and the same in Eq.~\ref{mix_loss}.
%\begin{center}
    \begin{equation}
    r_{mix} = wr_{1}+(1-w)r_{2}
    \label{mix_ratio}
    \end{equation}
    \begin{equation}
        L^{CLIP} = min(r_{mix}\hat{A}_t,clip(r_{mix},1-\epsilon,1+\epsilon)\hat{A}_t)
    \label{mix_ratio_loss}
    \end{equation}
%\end{center}
Another option is to mix two different losses as Eq.~\ref{mix_loss}.  In this case, the sample would be invalid only if both ratios were out of range.
%\begin{center}
    \begin{equation}
        L^{CLIP}_{1} = min(r_{1}\hat{A}_t,clip(r_{1},1-\epsilon,1+\epsilon)\hat{A}_t)
    \end{equation}
    \begin{equation}
        L^{CLIP}_{2} = min(r_{2}\hat{A}_t,clip(r_{2},1-\epsilon,1+\epsilon)\hat{A}_t)
    \end{equation}
    \begin{equation}
        L^{CLIP} = wL^{CLIP}_{1}+(1-w)L^{CLIP}_{2}
    \label{mix_loss}
    \end{equation}
%\end{center}

\begin{algorithm}
\caption{PPO with mixed loss}  
\begin{algorithmic}
\STATE Initialize $\pi_{\theta}$
\FOR{iteration=1, 2, . . .}
\item run $\pi_{\theta}$ in environment, get experiences
\item Compute advantage estimate $\hat{A}_t$ with the experiences
\item $r_{1} = \frac{\prod\pi^{i}_{new}(a^{i}_{t}|s_{t})}{\prod\pi^{i}_{old}(a^{i}_{t}|s_{t})}$
\item $r^{i}_{2} = \frac{\pi^{i}_{new}(a^{i}_{t}|s^{i}_{t})}{\pi^{i}_{old}(a^{i}_{t}|s^{i}_{t})}$
\IF{mix ratio loss}
\item $r_{mix} = wr_{1}+(1-w)r_{2}$
\item $L^{CLIP} = min(r_{mix}\hat{A}_t,clip(r_{mix},1-\epsilon,1+\epsilon)\hat{A}_t)$
\ENDIF
\IF{mix loss}
\item $L^{CLIP}_{1} = min(r_{1}\hat{A}_t,clip(r_{1},1-\epsilon,1+\epsilon)\hat{A}_t)$
\item $ L^{CLIP}_{2} = min(r_{2}\hat{A}_t,clip(r_{2},1-\epsilon,1+\epsilon)\hat{A}_t)$
\item $L^{CLIP} = wL^{CLIP}_{1}+(1-w)L^{CLIP}_{2}$
\ENDIF
\item compute gradient
\item update $\pi_{\theta}$
\ENDFOR
\end{algorithmic}
\label{ppo_algo}
\end{algorithm}

\section{Experiments}
We validated our idea in Gym-$\mu$RTS~\cite{huang2021gym} and MuJoCo~\cite{todorov2012MuJoCo}. In our experiments, we will mainly investigate four different losses, as shown in Table \ref{tab:table1}. In Gym-$\mu$RTS, we used the winning rate of the last 100 games and the average reward of the last 100,000 steps in the training process as indicators of model performance. In MuJoCo, the total reward of each round is used as a measure of model performance. 

As far as the code is concerned, if the difference between the old policy and new policy is small, then the clip of PPO will not be applied. In a common serial architecture, reusing samples improves efficiency. The smaller sample re-use time and the number of samples in small batches may result in less clipping.

Considering the work of Engstrom et al.~\cite{engstrom2020implementation} on PPO, we did not use additional optimizations of the standard implementation of PPO in our code. And given that in the work of Engstrom et al.~\cite{engstrom2020implementation}, PPO without and with clipping have similar results, we conduct experiments in our implementation. In our implementation, PPO with clipping is far superior to PPO without clipping, which means that clipping is important as shown in our experiments in Figures \ref{fig:mujoco_clip_rewards} and Figures \ref{micrortsrewards}. The hyperparameters we used are basically the same as those in the PPO paper \cite{schulman2017proximal}.

\begin{table*}[t]
  \centering
  \begin{tabular}{l|l|l}
    Loss     & Equation     & Description \\
    compound action loss & (\ref{r1})  & Calculate the probability of the entire action (probability density)\\
    sub-action loss     & (\ref{r2}) & Calculate the probability of each sub-action (probability density)\\
    mix ratio loss     & (\ref{mix_ratio_loss})       & Calculate loss using weighted average of two ratios\\
    mix loss     & (\ref{mix_loss}) &Weighted average of compound action loss and sub-action loss\\
  \end{tabular}
  \caption{List of losses studied in this paper.}
  \label{tab:table1}
\end{table*}

We mainly use an asynchronous PPO architecture as Algorithm \ref{appo_algo}, where the sampler and trainer are two separate processes. This structure can be deployed on a large scale, where the new strategy can be updated asynchronously. Using this structure, there will be a large gap between the old strategy when sampling and the new strategy when updating and clipping will play a role in such an architecture. In the experiments in MuJoCo we also use a serial OPP same as OpenAI baseline~\cite{baselines} for supplement as shown in Figure\ref{ppo_algo}. The random seed for the experiment shown in this paper is set to 0. In the experiments, we did not use any GPU. We use an Intel(R) Xeon(R) CPU E5-2650 v4 @ 2.20GHz for our experiments.

\begin{figure}[t]
    \includegraphics[width=0.9\textwidth]{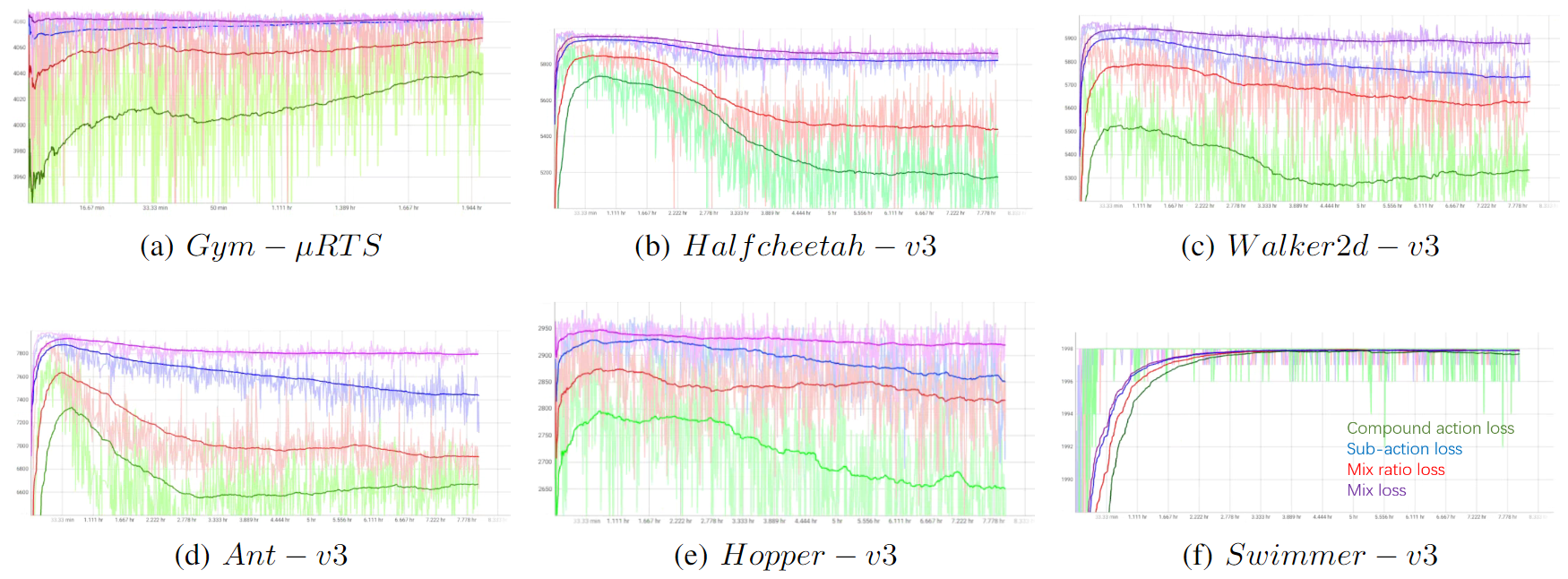}
    \caption{Number of unclipped samples in different environments during training.}
    \label{unclip_number}
\end{figure}

Before testing the effectiveness of our proposed method, we conducted a pre-experiment to count the number of unclipped samples with different losses to verify our ideas in Section 3. The hyper-parameters of this experiment are the same as those in Section 4.1 and 4.2. The details are shown in Figure \ref{unclip_number}. The experimental results confirm that in training, when the compound action loss is used, the number of samples that are not clipped is the least; when the sub-action loss is used, the number of samples that are not clipped is larger; when the mix ratio loss is used, the number of samples that are not clipped is between two between them; and when using the mix loss, the number of samples that are not clipped is the largest, because it is a combination of the first two (the weight of two losses is 0.5, the influence of a sample will be less then in the other loss). 

\begin{algorithm}
\caption{Asynchronous PPO}  
\begin{algorithmic}
\STATE Sampler process:
\STATE Initialize $\pi_{sampler}$, exp buffer
\FOR{iteration=1, 2, . . .}
\item copy neural network parameters $\theta$ from database to $\pi_{sampler}$
\item run $\pi_{sampler}$ in environment, get experiences and send to database
\ENDFOR
\STATE Trainer process:
\STATE Initialize $\pi_{train}$
\FOR{iteration=1, 2, . . .}
\item get experiences from database
\item Compute advantage estimate $\hat{A}_t$
\item Compute gradient 
\item update $\pi_{train}$
\item copy $\theta$ of $\pi_{train}$ to database
\ENDFOR
\end{algorithmic}
\label{appo_algo}
\end{algorithm}

\subsection{Experiments in MuJoCo}
MuJoCo is a physics engine designed to facilitate research and development in different fields that require fast and accurate simulations. We conduct experiments in the MuJoCo environment made by OpenAI~\cite{mujocodoc}. We use Version 1.31 of MuJoCo (distributed with an MIT License). MuJoCo's reward consists of three parts: the forward speed, action consumption and survival reward. That is, in this environment, the robot should move forward as quickly as possible and consume as little energy as possible. The proportion of the three rewards is different in different environments. The observation space and action space of the five MuJoCo environments in this experiment are shown in Table \ref{tab:MuJoCo enviroment}. Among them, the action is the torque applied to each joint of the robot, and observations include the position and speed of each part of the robot.

\begin{table*}[t]
  \centering
  \begin{tabular}{l|l|l|l}
    Environment & Observation Dim & Action Dim & Type\\
    Ant-v3 & 111 & 8 & continuous \\
    HalfCheetah-v3 & 17 & 6 & continuous\\
    Swimmer-v3 & 8 & 2 & continuous\\
    Hopper-v3 & 11 & 3 & continuous \\
    Walker2d-v3 & 17 & 6 & continuous\\
    Humanoid-v3 & 376 & 17 &continuous\\
    Gym-$\mu$RTS 10x10 map& (10,10,27) & 8 & discrete\\
    Gym-$\mu$RTS 16x16 map& (16,16,27) & 8 & discrete\\
  \end{tabular}
  \caption{Observation space and action space of different MuJoCo and Gym-$\mu$RTS environment}
  \label{tab:MuJoCo enviroment}
\end{table*}

     In the experiments of MuJoCo, the discount factor $\gamma$ is 0.99, $\lambda$ for the GAE is 0.95, clip $\epsilon$ is 0.2, the entropy regularization coefficient is 0.001, the value coefficient is 1, the learning rate of the optimizer is 2.5e-4, and the training time is 8 hours. To represent this policy, we use a 3 layer fully connected MLP with a 64 unit hidden layer, and there are only two additional noise layers after the MLP for exploration during training. We do not share parameters between policy and value function. We used asynchronous PPO\ref{appo_algo} and serial PPO\ref{ppo_algo}.
 
\begin{figure}[t]
    \centering
    \includegraphics[width=0.9\textwidth]{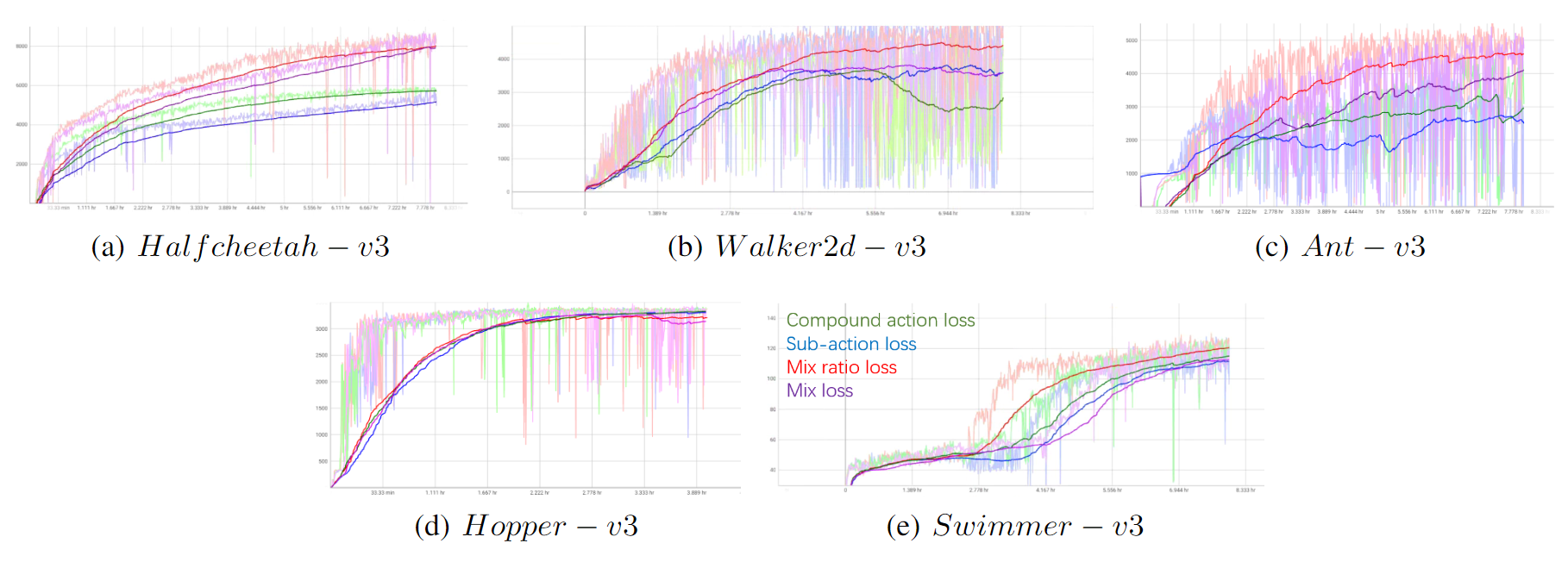}
    \caption{The performance of four losses during training in different MuJoCo environments. The mix ratio loss had the highest performance among Halfcheetah-v3, Walker2d-v3, and Ant-v3. Hopper-v3 and Swimmer-v3 have less space for action, Hopper-v3 has only 3 sub-actions and Swimmer-v3 has only 2 sub-actions. Their composite action loss and sub-action loss effects are similar, so the performance of the four losses in Hopper-3 and Swimmer-v3 is similar.}
    \label{fig:mujoco_m_rewards}
\end{figure}

We compare the experimental results with the benchmark of OpenAI, as shown in Table \ref{tab:table3}. Table \ref{tab:table3} records the average rewards of 100 episodes of four loss models after 8 hours of training in different MuJoCo environments and the benchmark of OpenAI.

\begin{table*}[t]
  \centering
  \begin{tabular}{l|l|l|l|l|l}
    Loss     & compound action & sub-action & mix ratio & mix loss& baseline\\
    Ant-v3 & 2955[2525,3366]  & 2495[1711,2649] & 4585[4480,5107] & 4099[3702,4547] & -\\
    HalfCheetah-v3 & 5166[4876,5332] & 5734[5535,6051] & 8065[7499,8224] & 7931[7766,8479] & 1668\\
    Swimmer-v3 & 109[106,116] & 108[106,116] & 108[100, 110] & 105[102,111] & 111\\
    Hopper-v3 & 3290[3065,3361] & 3166[3024,3309] & 3071[2984,3260] & 3480[3338,3651] & 2316\\
    Walker2d-v3 & 2846[2685,2941] & 3608[3402,3726] & 4415[4123,4521] & 3583[3386,3708] & 3424\\
  \end{tabular}
  \caption{Average rewards after training, 95\% confidence intervals computed from 100 trajectory of trained model. Baseline data is from OpenAI Github web \cite{baselines}.}
  \label{tab:table3}
\end{table*}

As a common loss, the compound action loss has a good performance in our implementation (Figure \ref{fig:mujoco_m_rewards} and Table\ref{tab:table3}). Compared with the experimental benchmark of OpenAI, the average episode reward after training is higher. And the performance of the mix ratio loss has a higher score of more than $50\%$ in Ant-v3 and Halfcheetah-v3 compared with the loss of compound action. Although the effect of using only the sub-action loss underperforms, the simple mixed loss combining it and the compound action loss can achieve a higher reward after training. The mix ratio loss received higher rewards in more experiments than the mix loss. In most of our experiments, the reward of the training curve using the sub-action loss is the lowest. We think this is because the sub-action loss does not consider the strategy of the whole action. The mixed loss is simply a combination of sub-action loss and compound action loss, that is, the learning rate is halved and using these two losses at the same time. The performance of this loss is also good. In the Ant-v3 and Halfcheetah-v3 environments, its reward is second only to the mix ratio loss. This experiment in different MuJoCo environments verifies the effectiveness of our proposed loss. 

\begin{figure}[t]
    \centering
    \includegraphics[width=0.9\textwidth]{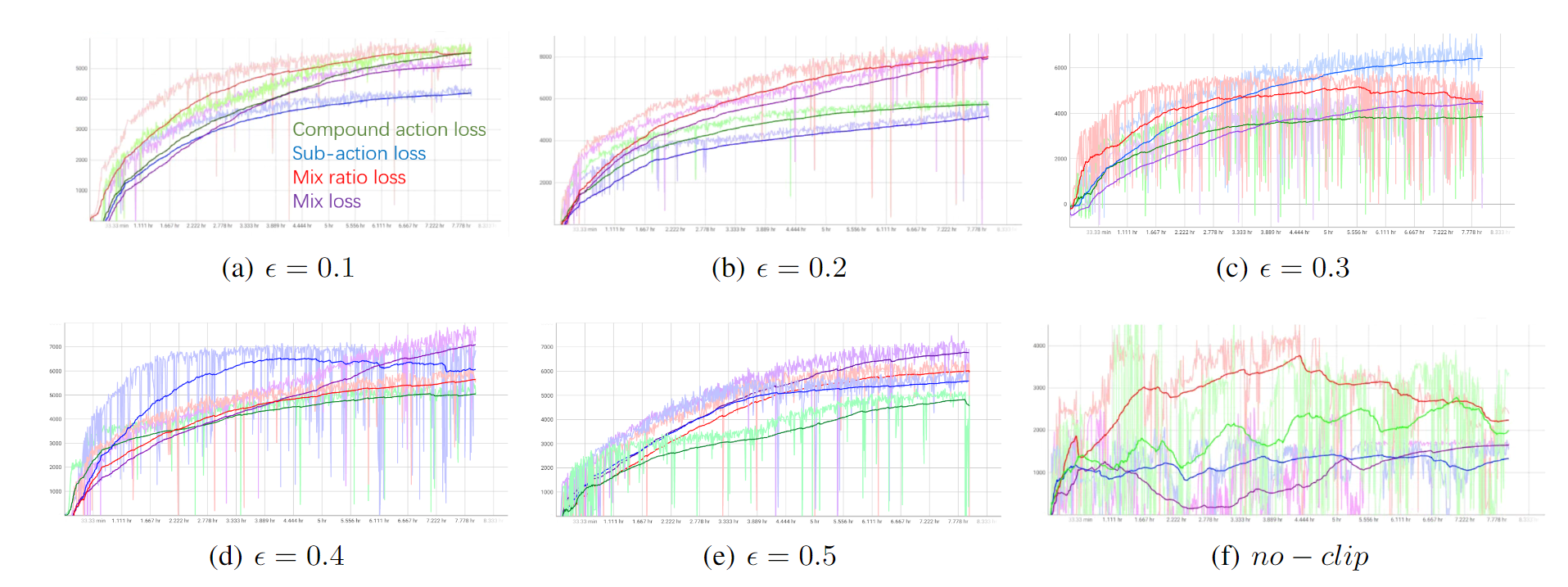}
    \caption{Performance in Halfcheetah-v3 during training with different clipping coefficient $\epsilon$. }
    \label{fig:mujoco_clip_rewards}
\end{figure}

As shown in Equation \ref{lclip}, the clip coefficient is the main coefficient of the loss in PPO, so we investigate the influence of this parameter. We carried out experiments under different clipping coefficients including no clipping. Although the pass rate of samples can be increased by using sub action loss and increasing the clip range, the sub action loss will not use samples that will lead to overestimated for updating because of increasing the clip range. Then we compare different losses in Halfcheetah-v3 under different clipping coefficients. The training curve of the ratio-mix-loss has a faster growth rate and a final reward when the clipping coefficient is small (as shown in Figure \ref{fig:mujoco_clip_rewards} (a) (b)). And the sub-action loss and the mix-loss have higher rewards than the compound action loss when the clip coefficient is large (as shown in Figure \ref{fig:mujoco_clip_rewards} (c) (d) (e)). In the case of no-clip, the training curves of four different loss types are difficult to converge.

At the same time, we also use the common serial PPO algorithm to carry out experiments. In this experiment, we applied standard methods to achieve stable training. These include: advanced normalization, gradient clipping, value clipping, state normalization, and reward scaling. If these methods are not used, performance curve will struggle to rise or will drop during training.

\begin{figure}[t]
    \centering
    \includegraphics[width=0.9\textwidth]{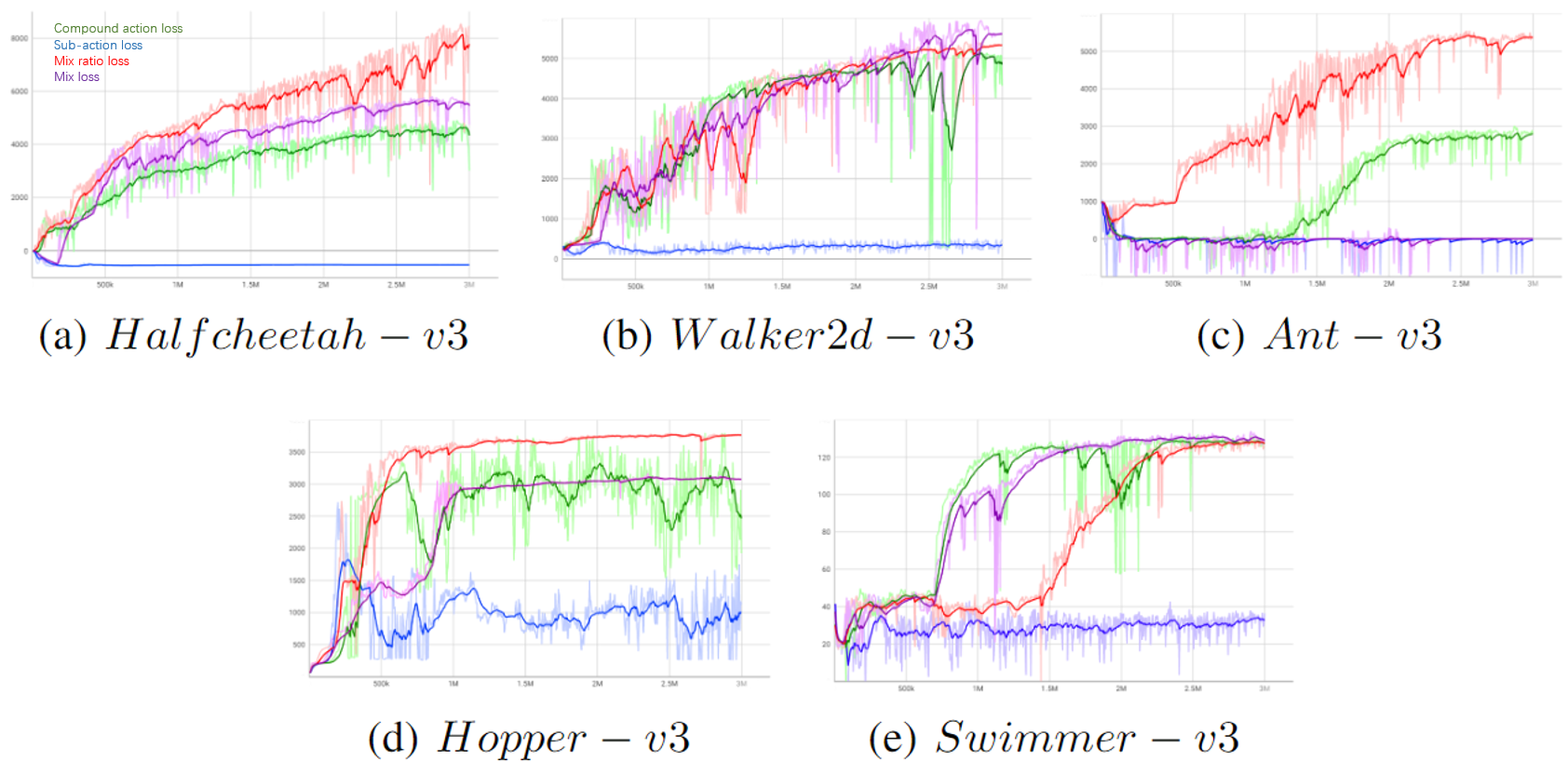}
    \caption{The performance of four losses during training in different MuJoCo environments with serial PPO algorithm. }
    \label{fig:mujoco_single_process_rewards}
\end{figure}

In this experiment, the model sampled 3M steps. The experimental results are shown in Figure \ref{fig:mujoco_single_process_rewards}. Similar with the experiments with asynchronous PPO, the mix-ratio-loss still performs well with the serial PPO algorithm. It has the best performance in Halfcheetah-v3, Ant-v3 and Hopper-v3 compared with other losses.  In the experiments with asynchronous PPO, the performance of the sub-action-loss and compound-action-loss are roughly the same. On the other hand, in the experiments with serial PPO, the performance curve of the sub-action-loss does not rise. The causes of the different performance of sub-action-loss in two experiment may be that in the asynchronous implementation more actors are used and there are more adequate sampling. And the mini-batch size of asynchronous PPO is larger than mini-batch size of serial PPO, which leads to more accurate gradients during updating.

\begin{figure}[t]
    \centering
    \includegraphics[width=0.9\textwidth]{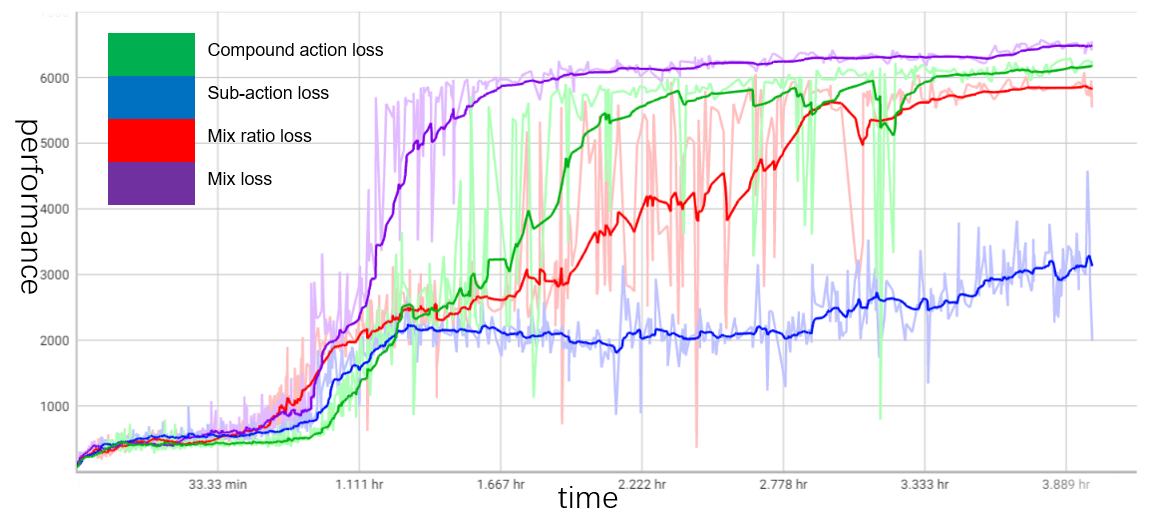}
    \caption{The training curve of four losses in Humanoid-v3.}
    \label{fig:humanoid result}
\end{figure}

In addition, we have an experiment in Humanoid-v3, which is regarded as a complex high-dimensional continuous control problem. The result shown in Figure~\ref{fig:humanoid result} is similar to results previously presented in this paper. And it is worth noting that sub-action-loss has significantly worse performance while mix-loss has best performance. 

Based on the previous experiment, it can be seen that in the MuJoCo environment, the sub-action loss alone does not work well, while the mixed loss (mix rtaio loss and mix loss) combining the sub-action loss and compound action loss results in better model training. As mentioned above, the sub-action-loss increases the pass rate of the samples and optimizes the strategy at each sub-action level, ignoring the optimization at the action combination level, which will bias the direction of the strategy optimization. The mixed loss makes the optimization more stable and obtains better training results while increasing sample utilization.

In the next section of this paper, we will conduct experiments in the Gym-$\mu$RTS environment, to investigate the performance of the different losses in discrete action environment.

\subsection{Experiments in Gym-microRTS}
 $\mu$RTS \cite{micrortsgithub} is a small implementation of an RTS game. Its action space and observation space are relatively small (compared to some complex environments such as StarCraft2), which allows agents to be trained with less resources and time. Gym-$\mu$RTS is the Python version of $\mu$RTS, which is based on OpenAI gym. Gym-$\mu$RTS provides different AI opponents, which can be used for testing and training. We use Version 0.3.2 of Gym-$\mu$RTS (GNU General Public License v3.0). The difference between Gym-$\mu$RTS and MuJoCo environment is that the actions in MuJoCo are continuous actions, while the actions in Gym-$\mu$RTS are discrete actions, the agent of Gym-$\mu$RTS must select a unit, decide what action to do and how to do it; MuJoCo will return reward at every step, whereas the reward of Gym-$\mu$RTS is sparse and related to game victory.

\begin{figure}[t]
    \centering
    \includegraphics[width=0.9\textwidth]{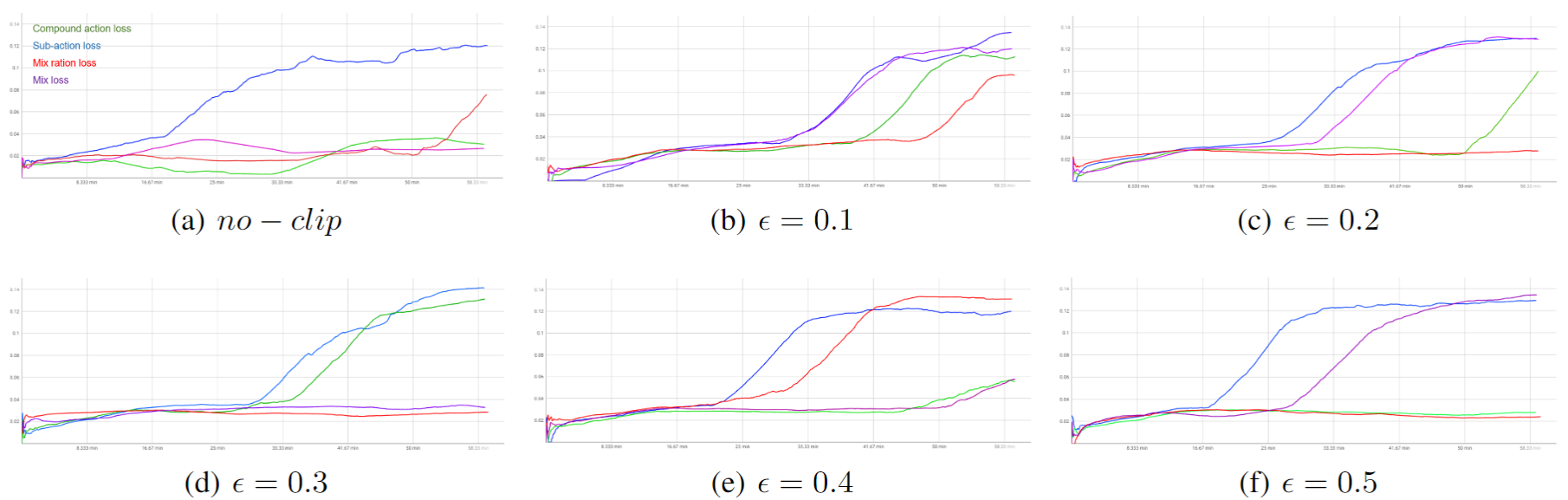}
    \caption{This figure show the average step reward of recent 100000 steps during training with different clipping coefficients $\epsilon$. In the Gym-$\mu$RTS environment, during the training process, the agent must explore to find a winning strategy, which will be optimized until the winning rate is 1. After the winning rate reaches 1, the winning efficiency will continue to be optimized and step reward will continue to increase slowly with the training.}
    \label{micrortsrewards}
\end{figure}

We experimented with different clipping ranges in the 10$\times$10 map against CoacAI who is the champion of the 2020 $\mu$RTS competition. In this experiment, the discount factor $\gamma$ is 0.99, $\lambda$ for the GAE is 0.95, the entropy regularization coefficient is 0.01, the value coefficient is 0.5, the steps number of a experience is 512, the number of environments is 32, the number of trainers is 3, and the iterations per epoch is 1. To represent this policy, we used the same neural network as Huang et al. \cite{huang2021gym}. The experimental results are shown in Figure \ref{micrortsrewards}.

\begin{figure}[t]
    \centering
    \includegraphics[width=0.9\textwidth]{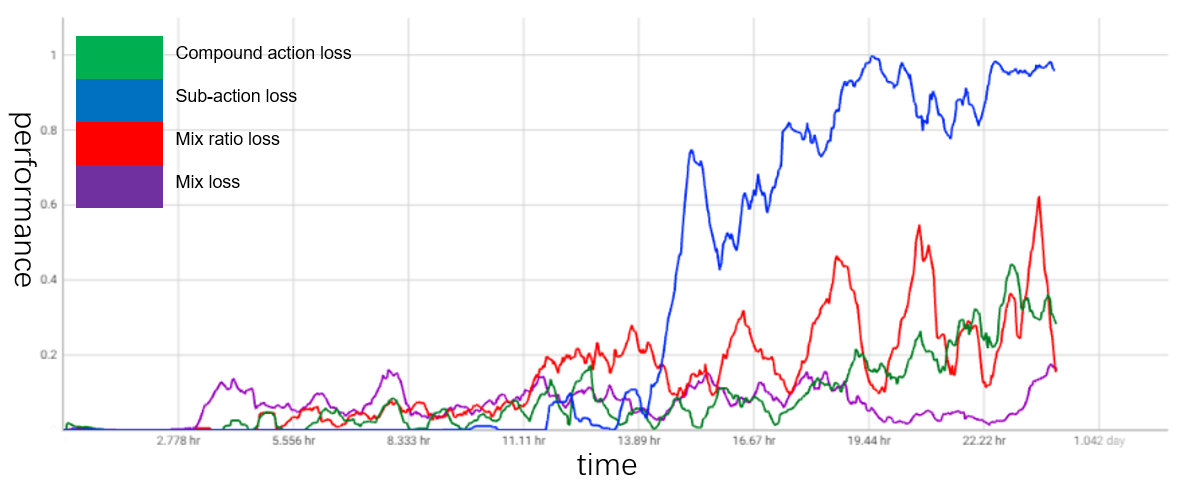}
    \caption{Winning rate of recent 1000 games against CoacAI during training in 16x16 map. Training model with sub-action loss can get nearly 100\% winning rate after 24h training.}
    \label{micrort16x16}
\end{figure}

In addition, we conducted experiments in a more complex 16x16 map. The training time of each experiment is 24 hours, with CoacAI as the opponent, and the winning rate of the last 100 games during training is used as the evaluation indicator. The experimental results are shown in Figure \ref{micrort16x16}. 

The experimental results in Gym-$\mu$RTS are different with the experiments in MuJoCo. Although the performance of PPO with the sub-action loss is not good in MuJoCo environments, it performs well in in Gym-$\mu$RTS, more quickly achieving a higher winning rate. When the range of clip increases, the training speed of PPO using the other three kinds of loss slows down and it is difficult to achieve a 100\% winning rate within one hour of training. The reason for this result may be related to the action of the agent. In the Gym-$\mu$RTS environment, the correlation between sub actions is very high, which is different from the assumption that sub actions are independent of each other by compound action loss. In Gym-$\mu$RTS environment, when an agent executes a compound action, it first needs to select a unit, then select the action type to be executed by the unit based on the selected unit, and then determine how to execute the action. This means that it is not appropriate to calculate the joint probability of sub actions by multiplying in the compound action loss. For the sub action loss, each sub action is updated separately, which reduces the correlation between sub actions and the update of sub action strategies that should not have been updated. 

\section{Conclusion and Discussion}
In conclusion, in this paper, we propose new loss functions for the PPO algorithm. We have shown that using the mixed loss of the sub-action and compound action can increase the efficiency of using sampled data, and take into account the optimization of compound action and sub-action at the same time, so as to obtain better training. Our experiments compare the training effects of the compound action loss, sub-action loss, mix ratio loss and mixed loss in Gym-$\mu$RTS and MuJoCo. In different experimental environments for continuous actions, the mix ratio loss can be used to obtain higher rewards in most cases. And in Gym-$\mu$RTS, the sub-action loss can be used to increase performance. We think it is better to increase the proportion of sub action loss in the environment with high correlation of sub actions. In our future work, we will explore more complex games. In addition, the weight of mixing of the two losses in this paper is the same. In fact, with the change of hyperparameters and the influence of the training environment, the weights should also change to achieve the optimal effect. Which loss should be weighted under what circumstances is a point worthy of further study, as is finding a better hybrid functional form instead of simply adding the weighted sum of ratio or loss. We suggest that if the correlation between sub actions is small, the weight of compound action ratio in mixed loss can be increased; otherwise, the weight of sub action ratio can be increased. Moreover, what causes the differences in the performance of different loss in different environments is also a problem worthy of in-depth study.

\end{document}